%% file: main.tex
\documentclass{article}

\PassOptionsToPackage{numbers, square,compress}{natbib}

\usepackage[final]{neurips_2022_ml4ad}

\usepackage[utf8]{inputenc} 
\usepackage[T1]{fontenc}    
\usepackage{hyperref}       
\usepackage{url}            
\usepackage{booktabs}       
\usepackage{multirow}
\usepackage{amsfonts}       
\usepackage{amssymb}
\usepackage{ marvosym }
\usepackage{nicefrac}       
\usepackage{microtype}      
\usepackage{xcolor}         
\usepackage{dsfont}
\usepackage{subcaption}
\usepackage{pifont}
\newcommand{\cmark}{\ding{51}}%
\newcommand{\xmark}{\ding{55}}%

\usepackage{floatrow}
\usepackage[pdftex]{graphicx}
\usepackage[noabbrev]{cleveref}

\DeclareRobustCommand{\abbrevcrefs}{%
\crefname{figure}{fig.}{figs.}%
\crefname{equation}{eqn.}{eqns.}%
\crefname{table}{tab.}{tabs.}%
}
\DeclareRobustCommand{\cshref}[1]{{\abbrevcrefs\cref{#1}}}

\DeclareRobustCommand{\todo}[1]{}

\renewcommand{\dagger}{\Cross}
\newcommand{\nc}{\textsuperscript{\dagger}}
\title{Analyzing Deep Learning Representations of Point Clouds for Real-Time In-Vehicle LiDAR Perception}

\author{%
  Marc Uecker\thanks{Corresponding author} \\
  FZI Research Center for\\
  Information Technology\\
  \texttt{uecker@fzi.de} \\
   \And
  Tobias Fleck\\
  FZI Research Center for\\ Information Technology\\
  \texttt{tobias.fleck@fzi.de} \\
  \And
  Marcel Pflugfelder \\
  Karlsruhe Institute of Technology \\
  \texttt{marcel.pflugfelder@student.kit.edu} \\
  \And
  J. Marius Zöllner\\
  Karlsruhe Institute of Technology, \\
  FZI Research Center for\\ Information Technology\\
  \texttt{zoellner@fzi.de} \\
}

\begin{document}%
\floatsetup[table]{capposition=top}%
\maketitle

\begin{abstract}
\input{Sections/00_Abstract}
\end{abstract}

\input{Sections/01_Introduction}

\input{Sections/02_Taxonomy}

\input{Sections/03_Analysis}

\input{Sections/04_Discussion}

\input{Sections/05_Conclusion}

\input{Sections/06_Acknowlegement}
\newpage
{
\small
\bibliography{bibliography.bib}
}





\end{document}

%% file: Sections/00_Abstract.tex
LiDAR sensors are an integral part of modern autonomous vehicles as they provide an accurate, high-resolution 3D representation of the vehicle's surroundings.
However, it is computationally difficult to make use of the ever-increasing amounts of data from multiple high-resolution LiDAR sensors.
As frame-rates, point cloud sizes and sensor resolutions increase, real-time processing of these point clouds must still extract semantics from this increasingly precise picture of the vehicle's environment.
One deciding factor of the run-time performance and accuracy of deep neural networks operating on these point clouds is the underlying data representation and the way it is computed.
In this work, we examine the relationship between the computational representations used in neural networks and their performance characteristics.
To this end, we propose a novel computational taxonomy of LiDAR point cloud representations used in modern deep neural networks for 3D point cloud processing.
Using this taxonomy, we perform a structured analysis of different families of approaches.
Thereby, we uncover common advantages and limitations in terms of computational efficiency, memory requirements, and representational capacity as measured by semantic segmentation performance.
Finally, we provide some insights and guidance for future developments in neural point cloud processing methods.

%% file: Sections/01_Introduction.tex
\section{Introduction}
\label{sec:intro}
Thanks to a large amount of investment into research from many stakeholders, the field of autonomous driving is progressing rapidly~\cite{yurtseverSurveyAutonomousDriving2020}.
One area where this is particularly evident is the field of LiDAR processing, which has recently been attracting increasing attention from the computer vision and deep learning community~\cite{guoDeepLearning3D2021}.
Meanwhile, the sensor hardware is also evolving.
As major players in the industry drive demand for lower-cost, high-resolution sensors, they are becoming more affordable and increasingly widespread~\cite{rorizAutomotiveLiDARTechnology2022}.
Following this trend, recent research vehicles and prototypes are often equipped with multiple high-resolution LiDAR sensors~\cite{wilsonArgoverseNextGeneration2021}.
Modern sensors are capable of delivering millions of points per second, at frame-rates at or above 10\,Hz for each sensor~\cite{carballoLIBREMultiple3d2020}.
\newpage
These fast and high-resolution sensors produce large amounts of data, which must be processed in real-time to be useful for the autonomous driving functions~\cite{strobelAccurateLowlatencyVisual2020}.
For many perception tasks which require semantic or geometric reasoning, such as object detection and semantic segmentation, only deep learning methods provide state-of-the-art processing capabilities~\cite{nguyen3DPointCloud2013,behleySemantickittiDatasetSemantic2019,xieLinkingPointsLabels2020,caesarNuscenesMultimodalDataset2020}.
However, many deep learning approaches which could be used to process LiDAR point clouds of such scale do not fulfill the real-time inference latency requirements for in-vehicle deployment~\cite{behleySemantickittiDatasetSemantic2019}.
We conjecture that the most important design decisions for inference run-time performance hinge on the underlying learned data representation.
Multiple papers categorize approaches as either point-based, projection-based or sometimes voxel-based to simplify comparison against state-of-the-art approaches~\cite{liuPVNAS3DNeural2021,xuRPVNetDeepEfficient2021,miliotoRangeNetFastAccurate2019}.
However, this categorization does not capture the full diversity of design decisions made in the development of new architectures.
We also found no substantive, objective analysis or comparison of the impact of these design decisions on run-time performance, as each paper focuses on the approach presented.\\
In this work, we present a taxonomy of different architecture designs, based on design decisions regarding the point cloud data representation.
We categorize approaches by their choice of explicit or implicit spatial structure, by their choice of internal representation dimensionality, their choice of coordinate space, and finally by their chosen method of feature aggregation.\\
This taxonomy is described in detail in \cref{sec:taxonomy}.
Using the introduced taxonomy, we analyze the impact of these design decisions on the run-time performance characteristics in \cref{sec:analysis}.
Finally, based on this analysis, we also provide insights and recommendations for future work in \cref{sec:discussion}.

%% file: Sections/02_Taxonomy.tex
\input{Sections/taxonomy_figure}

\section{Taxonomy of neural representations for LiDAR point clouds}
\label{sec:taxonomy}

In this section we describe our proposed taxonomy in detail.
The taxonomy is centered around the design decisions during development, which lead to a final representation of the point cloud inside a deep neural network. 
In \cref{fig:taxonomy} we illustrate the categorization of common pointcloud representations using our taxonomy. 
Notably, the categorization shown in \cref{fig:taxonomy} is not exhaustive, as there are many possible combinations of choices between the presented design decisions.

\paragraph{Spatial structure} The first design decision we observe (\cshref{fig:taxonomy}, first layer) is the choice between an explicit or implicit multi-dimensional spatial arrangement of data in memory.
An explicit spatial structure directly encodes positional information in the memory layout of the represented data.
Typically, a rasterized representation of a point cloud can be indexed by a point's coordinate to receive its feature vector.
In comparison, an implicit spatial structure stores the points' feature vectors in a sparse representation. 
In this case, the points' coordinates and/or a separate indexing data structure are often stored to encode positional information and accessed to extract neighborhood relations~\cite{huRandLANetEfficientSemantic2020,tangSearchingEfficient3D2020,thomasKPConvFlexibleDeformable2019}. 
\Cref{fig:explicit_implicit} visualizes the difference for an exemplary one-dimensional point cloud. 

\input{Sections/choices_figure_a}

\paragraph{Rasterization dimensionality} The second design decision we observe (\cshref{fig:taxonomy}, second layer) is the dimensionality of the internal mathematical representation of the point cloud.
The main varieties we observe are three-dimensional voxel representations, two-dimensional projections of 3D space, and one-dimensional unsorted set- or list-based representations~\cite{liuPointvoxelCNNEfficient2019,miliotoRangeNetFastAccurate2019,qiPointNetDeepHierarchical2017}.
We refer to the one-dimensional representations as "Bag-of-Points", as their order is typically irrelevant for the operations performed on them.
The multi-dimensional representations perform a rasterization of the space into a finite number of grid cells, aligning each memory cell to a section of 3D or 2D space.
The decision of representation dimensionality is orthogonal to the memory layout, as multi-dimensional rasterizations can also be sparsely stored in one-dimensional data structures~\cite{tangSearchingEfficient3D2020}.
\Cref{fig:dimensionality} illustrates different rasterization dimensionalities.

\paragraph{Coordinate system} The third design decision (\cshref{fig:taxonomy}, third layer) we observe concerns the choice of coordinate system, which is used for rasterizations of multi-dimensional spaces.
Rasterization divides 2D or 3D space into chunks of finite size. 
This partition is typically performed along regular intervals across coordinate axes.
Therefore, the coordinate axes chosen for this division also impacts how the resulting representation partitions 3D space.
Here, we mainly differentiate between Cartesian coordinate systems, which refer to absolute positions in 3D Euclidean space~\cite{xuRPVNetDeepEfficient2021}, and polar coordinate systems which refer to locations by combinations of angles and distance measurements~\cite{zhangPolarnetImprovedGrid2020,triessScanbasedSemanticSegmentation2020}.
\Cref{fig:cartesian_polar} (left) illustrates how different coordinate systems can lead to different rasterizations of two-dimensional space.\\
Spherical coordinates are an extension of polar coordinates which uses two angles and one distance measurement to index 3D space.
Projecting spherical coordinates along the radial axis results in a range image 2D representation.
There are also various coordinate systems which combine polar and Cartesian geometry for different axes.
\Cref{fig:cartesian_polar} shows an example of this: Cylinder coordinates use a polar coordinate system for the X-Y plane and a Cartesian axis for the Z-direction.
Similarly, some approaches use a polar coordinate system in a 2D Bird's eye view (BEV) projection, which projects points along a Cartesian z-axis~\cite{zhangPolarnetImprovedGrid2020}.
For polar coordinate systems, the coordinate origin is typically chosen as the center of the LiDAR sensor in order to minimize aliasing~\cite{zhangPolarnetImprovedGrid2020,triessScanbasedSemanticSegmentation2020}.

\paragraph{Feature aggregation} As a final design decision (\cshref{fig:taxonomy}, fourth layer), we differentiate approaches by their choice of mathematical operation to be applied to compute the resulting point cloud representation.
To compute a feature representation for a point in a 3D point cloud, almost all deep learning approaches aggregate information about its local or global neighborhood.
This aggregation typically requires finding other points within the neighborhood of the point whose features are to be computed.
Next, these features are aggregated using parametric or non-parametric mathematical operations.
The decision for a rasterized representation often directly leads to the use of convolutions~\cite{miliotoRangeNetFastAccurate2019,liuPVNAS3DNeural2021}, although other operations are certainly possible.
For Bag-of-Points representations, the choice of the feature aggregation method becomes the main differentiating factor.
Some approaches use convolutions in non-rasterized spaces~\cite{tatarchenkoTangentConvolutionsDense2018,thomasKPConvFlexibleDeformable2019,songDGPolarNetDynamicGraph2022}, others perform feature aggregation through different variants of weighted or non-weighted pooling of local neighbors~\cite{qiPointNetDeepLearning2017,huRandLANetEfficientSemantic2020,qiPointNetDeepHierarchical2017}.
For brevity, we group these into neighbor-based approaches in \cref{fig:taxonomy} (bottom right).
A final group of approaches which exist for smaller point clouds, but have not yet found application in the large-scale point clouds used in autonomous driving are point cloud transformers.
These approaches use local or global (point cloud-wide) attention mechanisms to exchange features of points inside a point cloud~\cite{luTransformers3DPoint2022}.

We note that there are likely additional possible options for each of the design decisions presented here, and encourage future work to experiment with possible alternatives.
We provide this taxonomy as a framework to analyze families of approaches used in practice.
In the following, we analyze each design decision from the taxonomy on its own, before continuing to a more high-level discussion of relevant design decisions for memory, run-time and representational capacity.

\input{Sections/choices_figure_b}

%% file: Sections/taxonomy_figure.tex
\begin{figure}[t!]
    \centering
    \includegraphics[width=\linewidth]{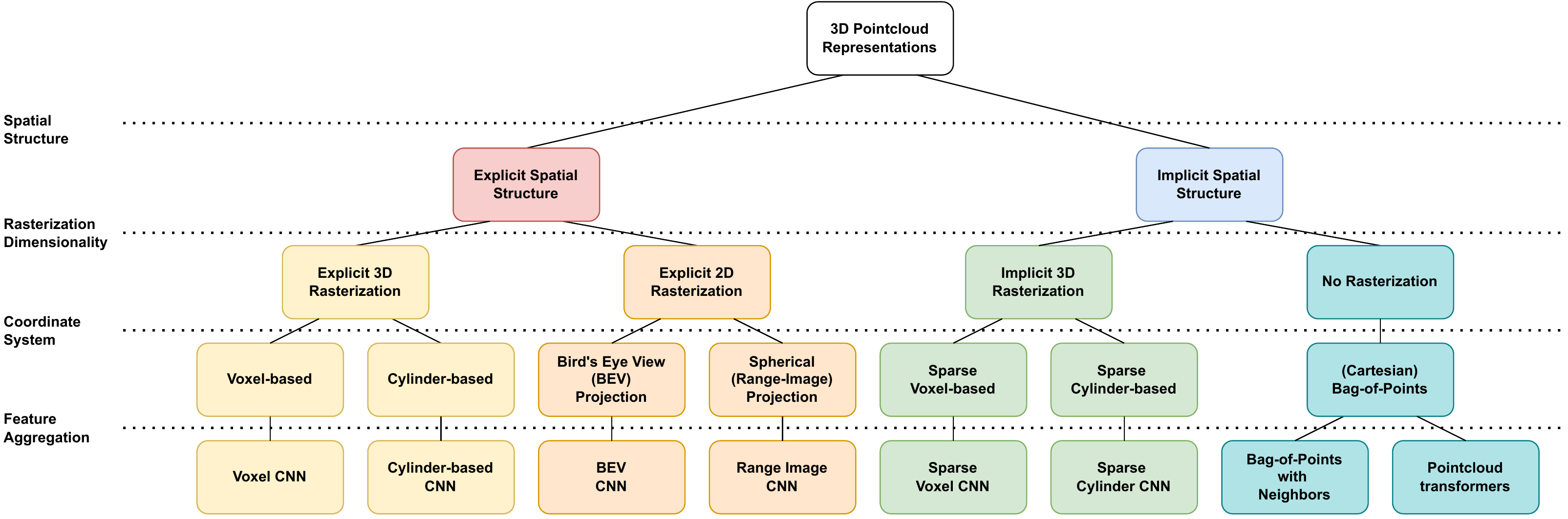}
    \caption{A categorization of common families of approaches using our proposed taxonomy for neural LiDAR pointcloud representations.}
    \label{fig:taxonomy}
\end{figure}

%% file: Sections/choices_figure_a.tex
\begin{figure}[t!]
\begin{floatrow}
\floatbox{figure}[0.5\Xhsize][\FBheight][t]{}{
    \includegraphics[width=\linewidth]{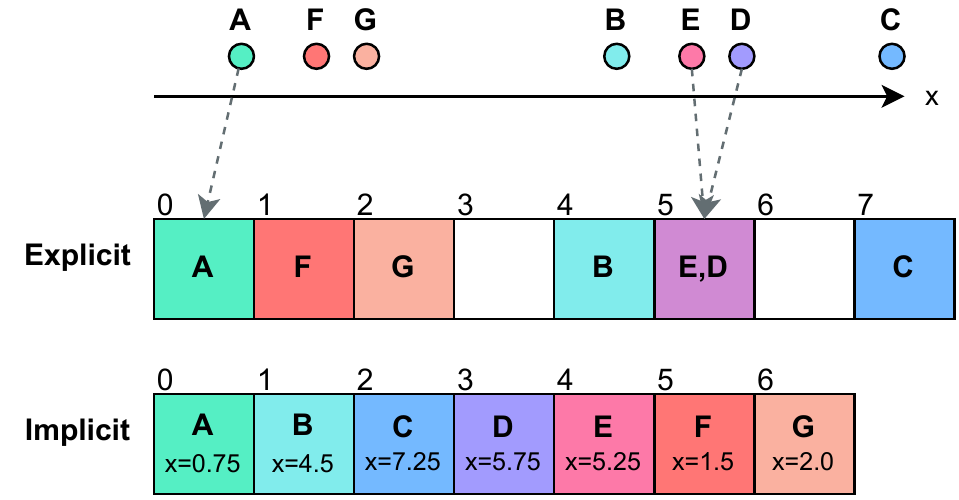}
    \caption{A comparison of explicit and implicit data representations. The example shows a 1-dimensional pointcloud on the top, and an explicit and implicit memory layout at the bottom.
    An example of aliasing can be observed, as points E and D collide into the same memory cell. 
    Wasted memory space can be seen as empty memory cells.}
    \label{fig:explicit_implicit}
}
\floatbox{figure}[\Xhsize][\FBheight][t]{}{
    \includegraphics[width=\linewidth]{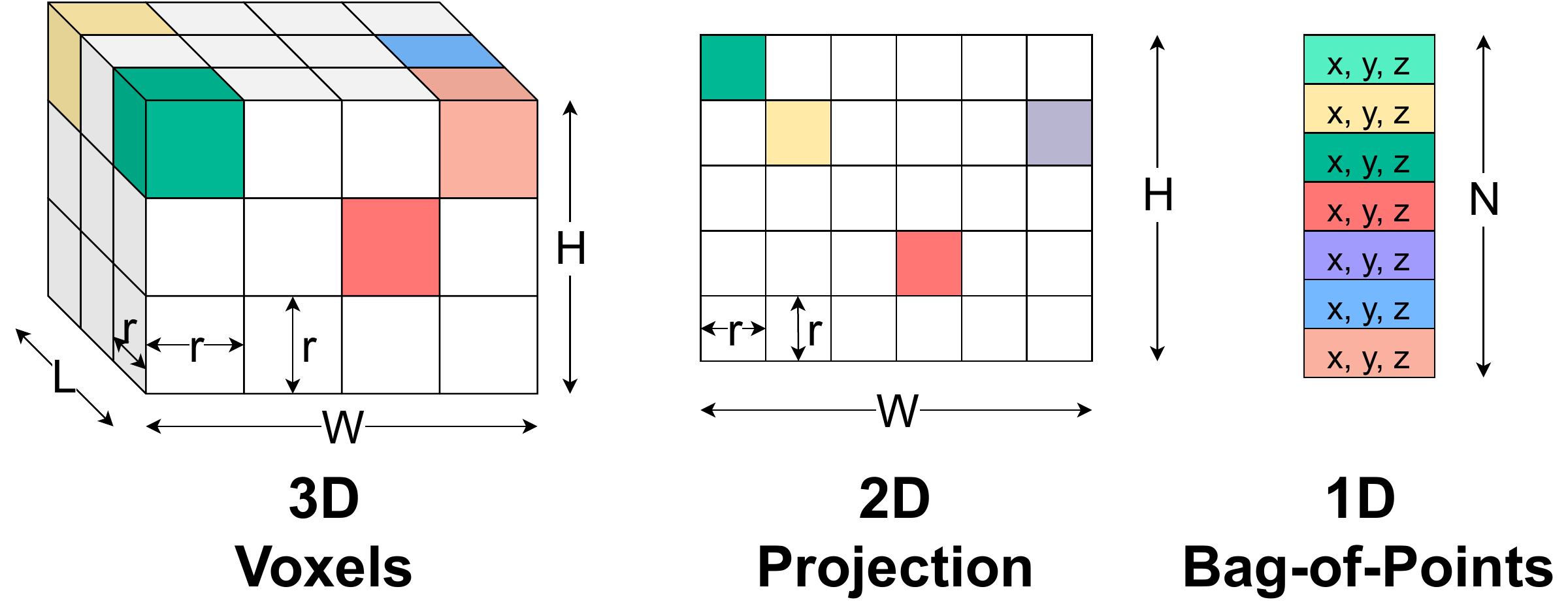}
    \caption{A comparison of 3D voxel representations, 2D Image representations and unordered 1-dimensional Bag-of-Points representations. As the resolution $r$ becomes more fine-grained, the sparsity of rasterized representations increases. However, a coarse resolution may cause multiple points to collide within a single representation cell (as seen in the top right of the 2D projection).}
    \label{fig:dimensionality}
}
\end{floatrow}
\end{figure}

%% file: Sections/choices_figure_b.tex
\begin{figure}[t!]
\begin{floatrow}
\TopFloatBoxes
\floatbox{figure}[0.5\Xhsize][\FBheight][t]{}{
\captionsetup{width=\linewidth}
\caption{Cartesian and polar coordinates shown in 2-Dimensional space. Cylinder coordinates are one way to combine both approaches in 3-dimensional space.}
\label{fig:cartesian_polar}
\includegraphics[width=\linewidth]{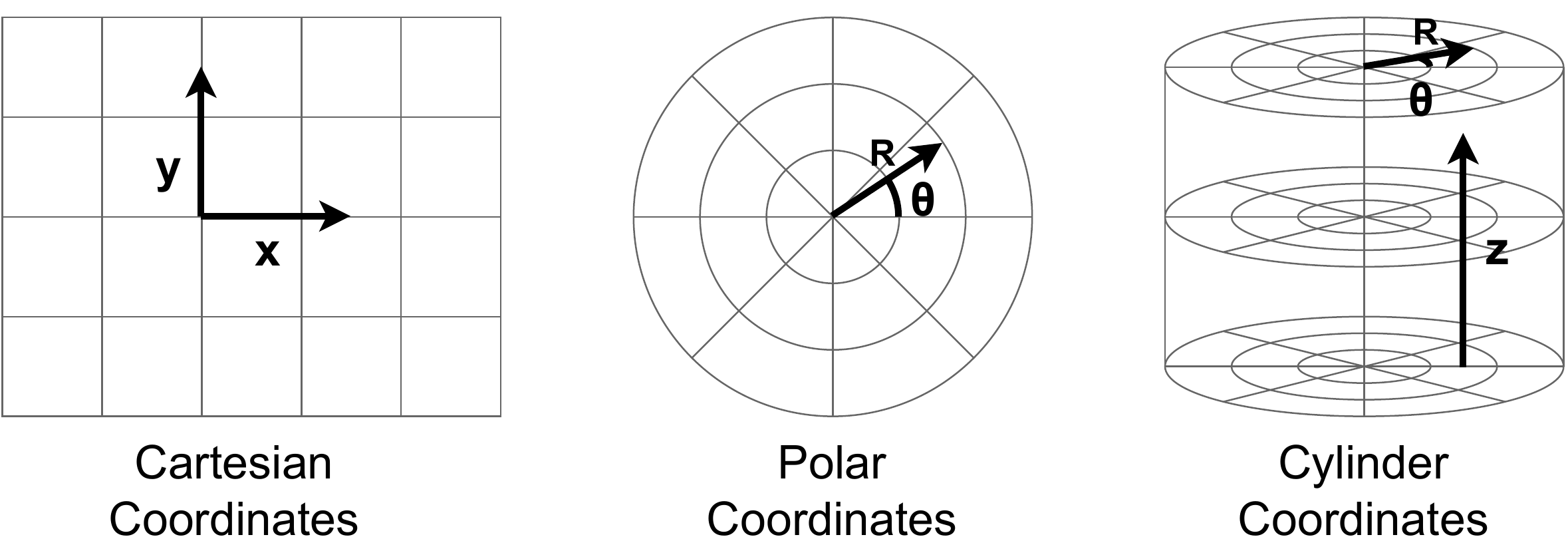}
}
\killfloatstyle
\floatbox[\captop]{table}[\Xhsize][\FBheight][t]{}{
\captionsetup{width=\linewidth}
\caption{Invariance properties of different representation families when combined with translation-invariant feature aggregation.\\
"Object rotation" refers to the rotation of an object around its center, whereas "Ego rotation" refers to a rotation of the sensor itself.}
\label{tab:cartesian_polar}
\resizebox{\linewidth}{!}{
\begin{tabular}{@{}lccc@{}}
\toprule
Representation Family & Translation & Ego rotation & Object rotation \\ \midrule
Range Image & \xmark & yaw, pitch & \xmark \\
Bird's Eye View (BEV) & \cmark & \xmark & \xmark \\
Polar BEV & z-axis & yaw & \xmark \\
3D Voxel & \cmark & \xmark &\xmark \\
3D Cylinder & z-axis & yaw & \xmark \\
(Cartesian) Bag-of-Points & \cmark & \xmark &\xmark \\
\bottomrule
\end{tabular}}}
\end{floatrow}
\end{figure}

%% file: Sections/03_Analysis.tex
\section{Analysis}
\label{sec:analysis}
In an autonomous vehicle, there are two major resource constraints which affect the perception system, namely memory and computation time.
The deployed hardware only provides a finite amount of memory.
Therefore, the amount of memory consumed by the point cloud representation should remain within reasonable scales.
More critically, computation time is severely limited in a moving vehicle, as any incurred latency makes a perception system's output outdated. 
There is a certain limit up to which an outdated observation remains useful~\cite{strobelAccurateLowlatencyVisual2020}.
Therefore, the latency should be as low as possible.
However, a more practical hard constraint is given by the rate that data is produced by the vehicle's sensors.
If a perception component's latency is less than the frame rate of the sensor whose data it processes (typically about 10 Hz for current LiDAR sensors~\cite{rorizAutomotiveLiDARTechnology2022}), the approach can be considered real-time capable~\cite{papadeasRealtimeSemanticImage2021}.

These resource constraints are contrasted by the representational capacity of a perception model, i.e. its ability to correctly infer information about the environment from the sensor data.
In this work, we use semantic segmentation performance as a proxy for representational capacity, based on the widely used SemanticKITTI\,\cite{behleySemantickittiDatasetSemantic2019} benchmark.
However, the same design decisions and implications presented here can likely be generalized to 3D object detection.
As a starting point for a similar analysis, multiple surveys provide an overview of recent 3D object detection methods~\cite{arnoldSurvey3DObject2019,qian3DObjectDetection2022,zamanakosComprehensiveSurveyLIDARbased2021}.\\
Typically, larger models have higher representational capacity, but also higher resource utilization~\cite{qiPointNetDeepHierarchical2017,liuConvnet2020s2022,liuPVNAS3DNeural2021}.
Therefore, the difficulty lies in creating a perception model with high representational capacity, which can still run real-time inference on in-vehicle hardware.

In the following sections, we examine each design decision from the taxonomy presented in \cref{sec:taxonomy} for its impact on memory consumption, run-time latency, and representational capacity.

\input{Sections/large_table}

\input{Sections/Analysis/01_explicit_implicit}

\input{Sections/Analysis/02_dimensionality}

\input{Sections/Analysis/03_coordinate_system}

\input{Sections/ops_scaling}

\input{Sections/Analysis/04_feature_aggregation}

%% file: Sections/large_table.tex
\begin{table}[t!]
\centering
\caption{A comparison of different approaches for 3D semantic segmentation, categorized by their used point cloud representation.
mIoU scores and Latency measurements are reported on the SemanticKITTI dataset.
The rightmost column provides the source for timing measurements.
An asterisk (*) denotes that the pre-processing time for projection (approx. 24 ms following \cite{liuPVNAS3DNeural2021}) was added to the number reported in the paper, as this is often omitted from latency comparisons for projection-based approaches~\cite{tangSearchingEfficient3D2020}. A dagger (\dagger) is used to mark approaches which have not published code, as they are not easily reproducible or usable for downstream tasks.}
\label{tab:comparison}
\resizebox{\textwidth}{!}{%
\begin{tabular}{@{}llrrrcl@{}}
\toprule
Method & Category in taxonomy & mIoU {[\%]} & Latency {[}ms{]} & \# params & Memory Complexity & Source \\ \midrule
PointNet\,\cite{qiPointNetDeepLearning2017} & Bag of Points (BoP) & 14.6 & 500 & 3M & $O(N\cdot d)$ & \cite{behleySemantickittiDatasetSemantic2019} \\ \midrule
PointNet++\,\cite{qiPointNetDeepHierarchical2017} & BoP + Neighbors & 20.1 & 5900 & 6M &\multirow{4}{*}{$O(N\cdot d)$} & \cite{behleySemantickittiDatasetSemantic2019} \\
TangentConv\,\cite{tatarchenkoTangentConvolutionsDense2018} & BoP + Neighbors & 40.9 & 3000 & 0.4M & & \cite{behleySemantickittiDatasetSemantic2019} \\
RandLA-Net\,\cite{huRandLANetEfficientSemantic2020} & BoP + Neighbors & 53.9 & 880 & 1.2M & & \cite{liuPVNAS3DNeural2021} \\
KPConv\,\cite{thomasKPConvFlexibleDeformable2019} & BoP + Neighbors & 58.8 & 279 & 18.3M & & \cite{liuPVNAS3DNeural2021} \\ \midrule
PVCNN\,\cite{liuPointvoxelCNNEfficient2019} & Voxel CNN & 39.0 & 146 & 2.5M & $O(\frac{L\cdot W \cdot H}{r^3})$ & \cite{liuPVNAS3DNeural2021} \\ \midrule
MinkowskiNet\,\cite{choy4DSpatiotemporalConvnets2019} & Sparse Voxel CNN & 63.1 & 294 & 21.7M & \multirow{5}{*}{$O(N\cdot d)$} & \cite{liuPVNAS3DNeural2021} \\
SPVCNN\,\cite{tangSearchingEfficient3D2020} & Sparse Voxel CNN & 58.8 & \textbf{85} & 21.8M & & \cite{liuPVNAS3DNeural2021} \\
SPVNAS\,\cite{tangSearchingEfficient3D2020} & Sparse Voxel CNN & 60.3 & \textbf{89} & 1.1M & & \cite{liuPVNAS3DNeural2021} \\
SPVNAS-A\,\cite{liuPVNAS3DNeural2021} & Sparse Voxel CNN & 63.7 & 110 & 2.6M & & \cite{liuPVNAS3DNeural2021} \\
SPVNAS-B\,\cite{liuPVNAS3DNeural2021} & Sparse Voxel CNN & 66.4 & 259 & 12.5M & & \cite{liuPVNAS3DNeural2021} \\ \midrule
Cylinder3D\,\cite{zhouCylinder3DEffective3D2020} & Sparse Cylinder-based & 67.8 & 170 & &  $O(N\cdot d)$ & \cite{houPointtoVoxelKnowledgeDistillation2022} \\ \midrule
SqueezeSegV2\,\cite{wuSqueezeSegv2ImprovedModel2019} & Range Image CNN & 39.7 & *\textbf{44} & 0.9M & \multirow{7}{*}{$O(\frac{W\cdot H}{r^2} \cdot d)$} & \cite{miliotoRangeNetFastAccurate2019} \\
SqueezeSegV3-21\,\cite{xuSqueezesegv3SpatiallyadaptiveConvolution2020} & Range Image CNN & 51.6 & \textbf{97} & 9.4M & & \cite{tangSearchingEfficient3D2020} \\
SqueezeSegV3-53\,\cite{xuSqueezesegv3SpatiallyadaptiveConvolution2020} & Range Image CNN & 55.9 & 238 & 26.2M & & \cite{tangSearchingEfficient3D2020} \\
RangeNet53\,\cite{miliotoRangeNetFastAccurate2019} & Range Image CNN & 49.9 & 102 & 50.4M & & \cite{liuPVNAS3DNeural2021} \\
RangeNet53++\,\cite{miliotoRangeNetFastAccurate2019} & Range Image CNN & 52.2 & *126 & & & \cite{miliotoRangeNetFastAccurate2019} \\
SalsaNext\,\cite{cortinhalSalsaNextFastUncertaintyaware2020} & Range Image CNN & 59.5 & \textbf{71} & 6.7M & & \cite{liuPVNAS3DNeural2021} \\
Lite-HDSeg\nc\,\cite{razaniLiteHDSegLiDARSemantic2021} & Range Image CNN & 63.8 & *\textbf{74} & & & \cite{razaniLiteHDSegLiDARSemantic2021} \\ \midrule
PolarNet\,\cite{zhangPolarnetImprovedGrid2020} & Polar BEV CNN & 54.3 & \textbf{62} & 13.6M & $O(\frac{W\cdot H}{r^2} \cdot d)$ & \cite{zhangPolarnetImprovedGrid2020} \\ \midrule
RPVNet\nc\,\cite{xuRPVNetDeepEfficient2021} & BoP + Voxel + Range Image CNN& 68.3 & 111 & 24.8M & \multirow{2}{*}{$O((N + \frac{W\cdot H}{r^2}) \cdot d)$} & \cite{xuRPVNetDeepEfficient2021} \\
CPGNet\nc\,\cite{liCPGNetCascadePointGrid2022} & BoP + BEV + Range Image CNN & 68.3 & *\textbf{91} & & & \cite{liCPGNetCascadePointGrid2022} \\ \bottomrule
\end{tabular}%
}
\end{table}

%% file: Sections/Analysis/01_explicit_implicit.tex
\subsection{Choice of spatial structure (explicit vs implicit)}
\label{sec:explicit_implicit}
In this section, we analyze the impact of the choice between an explicit or implicit spatial structure in the memory alignment of the point cloud representation.
The choice of spatial structure does not have a direct impact on representational capacity, as the same mathematical operations can be performed on both memory layouts.
Therefore, from a mathematical point of view, this choice is an implementation detail.
However, in practice, the memory layout can have a large impact on the run-time and memory consumption of a deep learning model.

One main advantage of an explicit spatial structure is that it directly encodes spatial information and therefore spatial locality through memory layout.
As a result, finding the feature vector for a particular location in the local neighborhood of a point can run in $O(1)$ time complexity regardless of point cloud size.
This is illustrated in \cref{fig:explicit_implicit}, where the neighborhood of point F in cell 1 can be accessed directly by accessing cells 0 and 2 in the explicit representation (center row).
In comparison, to find the neighborhood of point F in the implicit representation (bottom row), one must either traverse the entire list of $N$ points (an $O(N)$ operation), or construct and navigate some external indexing structure, such as a KD-tree~\cite{thomasKPConvFlexibleDeformable2019} or hashmap~\cite{tangSearchingEfficient3D2020}.
We list this neighbor lookup time as $t_\mathrm{lookup}$ in \cref{tab:repr_scaling}.
As an additional benefit, spatial locality in the memory layout often leads to better cache locality, as neighboring data is often already cached when its neighborhood has been accessed~\cite{liuPointvoxelCNNEfficient2019}.
On GPU accelerators, which are typically used for these models, accessing neighboring memory cells in parallel can also be much faster than random accesses due to architectural details~\cite{gutierrezMemoryLocalityExploitation2008,liuDataLayoutOptimization2013}.

The main disadvantage of an explicit spatial structure is given through the spatial sparsity of LiDAR data itself.
As the LiDAR points are not evenly distributed in 3D space, an explicit representation often incurs a significant fraction of empty memory cells, which represent sections of space with no LiDAR points.
Depending on the choice of data representation, a large percentage of memory can be occupied by empty spatial cells, where no LiDAR points reside~\cite{liuPointvoxelCNNEfficient2019,miliotoRangeNetFastAccurate2019}.
By default, implementations of operations such as convolutions do not treat these empty cells any differently than populated cells.
Therefore, not only memory, but also computation time is wasted on empty space.
This problem of wasted memory and computation cycles grows very quickly as the resolution $r$ of grid cells becomes finer, as can be inferred from \cref{fig:dimensionality}.
For 3D voxel-based methods, the number of spatial locations (i.e. voxels) $M$ in the representation, which dictates memory and run-time requirements, grows in the order of $O(\frac{L \cdot W \cdot H}{r^3})$. 
Here, $r$ is the side-length of voxels, and $L$, $W$, and $H$ are the side-lengths of the total volume to be encoded.
For 2D pixel-based methods, this scaling is similarly $O(\frac{W \cdot H}{r^2})$, where $r$ is the pixel size, and $W$ and $H$ are again the side-lengths of the 2-dimensional spatial region to be represented.
As the resolution $r$ tends to zero, the memory cost and run-time of both approaches grow rapidly.
Particularly for explicit 3D voxel approaches, the memory requirements tend to explode very quickly, limiting the practically usable resolution significantly\,\cite{liuPointvoxelCNNEfficient2019,liuPVNAS3DNeural2021}.

In summary, as deep neural networks perform aggregation of local neighborhoods of points, the choice of explicit or implicit representation is currently a trade-off between increasing either the number of locations $M$ whose local context is aggregated, or the cost of neighborhood queries $t_\mathrm{lookup}$, and therefore neighborhood aggregation for an individual location.
We believe future work could de-correlate the memory and time costs of an explicit representation, by operating sparsely on an explicit memory layout.
This would bring both the advantage of spatial locality as well as fast neighborhood queries, at the cost of large memory overhead.

%% file: Sections/Analysis/02_dimensionality.tex
\subsection{Choice of rasterization dimensionality (3D voxels vs 2D image vs 1D Bag-of-Points)}
\label{sec:dimensionality}

The choice of dimensionality for the internal -- mathematical -- point cloud representation is very important for both run-time and representational capacity.
If an explicit representation is chosen, the rasterization dimensionality of the internal representation also affects the scaling of memory costs, as explained in \cref{sec:explicit_implicit}.

Constructing and accessing both rasterized and non-rasterized representations takes a non-negligible amount of time, especially in a low-latency use case.
The construction and access times of representations depend significantly on the rasterization dimensionality, which we compare qualitatively in \cref{tab:repr_scaling}.\\
As described in \cref{sec:taxonomy}, deep neural networks on point clouds typically aggregate neighborhood information to infer the features of a point.
Depending on whether the representation is rasterized (2D or 3D) or not (Bag of Points), finding neighbors can either be a lookup in a data structure, or a nearest-neighbor search, which significantly affects the neighbor lookup latency $t_\mathrm{lookup}$.
To reduce neighborhood lookup times, Bag-of-Points representations are typically combined with search tree index structures such as KD-trees~\cite{thomasKPConvFlexibleDeformable2019}.
However, even in these tree structures, neighborhood search remains expensive, and their construction takes significant time as well.
As a result, non-rasterized representations (BoP + Neighbors in \cshref{tab:comparison}) typically take much longer to compute and process their internal representation than rasterized methods (range image or sparse voxel CNN in \cshref{tab:comparison}) for similar numbers of parameters and mIoU scores~\cite{xuRPVNetDeepEfficient2021}.
While accessing rasterized representations is much faster, their construction takes a non-negligible amount of time.
For 2D projection-based approaches, Liu et al.~\cite{liuPVNAS3DNeural2021} report approximately 24~milliseconds of latency for the projection computation alone\footnote{We suspect this projection time was measured on the CPU and GPU implementations may be faster. However, it is the only source we found for this projection latency.}.
For sparse 3D rasterizations, this equates to a hashmap insertion for each point~\cite{tangSearchingEfficient3D2020}.

Due to these run-time constraints, sparse voxel-based approaches tend to have less than ten million parameters (15-20 billion multiply-accumulate (MAC) operations as reported in \cite{tangSearchingEfficient3D2020}) before exceeding 100ms of latency.
In comparison, 2D CNNs with an explicit representation have a much higher limit of tens of millions of parameters (approximately 200 billion MACs according to \cite{tangSearchingEfficient3D2020}).
Therefore, sparse 3D convolutional neural networks appear to have much lower parameter budgets than 2D CNNs for similar latency constraints.
However, the IoU scores for these 3D models are typically higher than 2D CNNs with comparable numbers of parameters (compare range image vs sparse voxel CNNs in \cshref{tab:comparison}).
This indicates that the 3D representation enables a higher representational capacity for similar parameter budgets.
This is supported by an analysis performed by Triess et al. \cite{triessScanbasedSemanticSegmentation2020}, who find that 2D Range image based CNNs quickly reach diminishing returns with regards to scaling parameter counts.
We conjecture that this may be due to inherent invariances present in varying representations, and discuss this point further in \cref{sec:cartesian_polar}.

As the dimensionality of a rasterized representation is changed, this also affects both the size and structure of a point's neighborhood.
In a 3D voxel representation, each voxel has up to six direct neighbors, and a neighborhood of 27 voxels is considered for a 3x3x3 convolution .
In 2D, each pixel has four direct neighbors, and a 3x3 convolution considers a neighborhood of 9 pixels.
The time required for a typical neighborhood aggregation operation scales with the number of accessed neighbors ($n_\mathrm{neighbors}$ in \cshref{tab:repr_scaling,tab:ops_scaling}), which is significantly affected by the rasterization dimensionality~\cite{tangSearchingEfficient3D2020}.
This effect on neighborhood size is an important consideration for inference time.

Another important consideration for the choice of rasterization dimensionality is the resulting sparsity of the representation.
A typical three-dimensional voxel space will be considerably more sparsely populated with points than a 2D range image projection~\cite{liuPointvoxelCNNEfficient2019,triessScanbasedSemanticSegmentation2020}.
This sparsity also affects the choice of explicit or implicit representation, as discussed in \cref{sec:explicit_implicit}.


%% file: Sections/Analysis/03_coordinate_system.tex
\subsection{Choice of coordinate system (Cartesian vs polar)}
\label{sec:cartesian_polar}
The choice of mathematical coordinate system is very important for representational capacity.
As the coordinate system imparts a notion of distance, this choice affects what is regarded as the local neighborhood of a point.
Since deep learning methods typically aggregate features over this local neighborhood, this also affects representation capacity and, to a lesser extent, run-time latency.

Given that almost all rasterization-based approaches use convolutions to aggregate features, they inherit translation invariance in their internal coordinate system.
Depending on this coordinate system, this may lead to different invariances in 3D space, which again affects representational capacity.
\Cref{tab:cartesian_polar} compares the invariances of various rasterized coordinate systems when utilized with a translation-invariant feature aggregation method.
There currently appears to exist a mutual exclusivity for translation invariance and invariance against rotations around the sensor.
Additionally, no listed rasterized representation provides rotation invariance around arbitrary axes.
Based on representational capacity, we find no clearly preferred coordinate system.
However, recent results indicate that utilizing multiple representations including coordinate systems with both rotational and translational invariances may provide noticeable improvements in semantic segmentation performance~\cite{xuRPVNetDeepEfficient2021,liCPGNetCascadePointGrid2022}.

Since LiDAR sensors scan the environment using radial beams, the density of the resulting point clouds decreases with increasing distance~\cite{carballoLIBREMultiple3d2020}.
In Cartesian coordinate systems, this leads to large differences in the local density of points, which may impede training stability.
Polar coordinate systems do not suffer from this imbalance, at the cost of utilizing non-uniform spatial sections, which may again negatively affect representational capacity.\\
Angle-based representations such as range image projections are also sensitive to the choice of coordinate origin.
This sensitivity makes their use in multi-sensor applications more challenging, as the points are no longer evenly distributed around a radial center.


%% file: Sections/ops_scaling.tex
\begin{table}[t!]
\begin{floatrow}
\BottomFloatBoxes
\floatbox[\captop]{table}[0.5\Xhsize][\FBheight][t]{}{%
\captionsetup{width=\linewidth}%
\caption{Effects of representation dimensionality and memory layout on run-time determining factors. Actual timings depend on implementation and hardware details.}%
\label{tab:repr_scaling}%
\resizebox{\linewidth}{!}{%
\begin{tabular}{@{}lcccc@{}}%
\toprule%
Representation & $M$ & construction time & $t_\mathrm{lookup}$ & $n_\mathrm{neighbors}$\\ \midrule%
Explicit 3D & very high &high & low  &  cubic \\
Explicit 2D & medium-high & medium & low & quadratic\\
Implicit 3D & low-medium &low-medium & medium & cubic \\
Bag-of-Points & low-medium &medium-high  & high & configurable \\ \bottomrule
\end{tabular}%
}}
\floatbox[\captop]{table}[\Xhsize][\FBheight][t]{}{
\captionsetup{width=\linewidth}
\caption{Run-time Latency and Parameter scaling of different feature aggregation operations.}
\label{tab:ops_scaling}
\resizebox{\linewidth}{!}{
\begin{tabular}{@{}llc@{}}
\toprule
Aggregation           & Latency & Parameters \\ \midrule
Convolution           & $O(M \cdot n_\mathrm{neighbors} \cdot (t_\mathrm{lookup} + d^2))$ &  $n_\mathrm{neighbors} \cdot d^2$          \\
Depth-wise Convolution & $O(M \cdot n_\mathrm{neighbors} \cdot (t_\mathrm{lookup} + d))$ & $n_\mathrm{neighbors} \cdot d$\\
Pooling               & $O(M \cdot n_\mathrm{neighbors} \cdot (t_\mathrm{lookup} + d))$ & $0$ \\
Attention             & $O(M \cdot n_\mathrm{neighbors} \cdot (t_\mathrm{lookup} + d))$ & $0$ \\ \midrule 
Fully-connected               & $O(M \cdot (t_\mathrm{lookup} + d^2))$ & $d^2$ \\
\bottomrule
\end{tabular}%
}}
\end{floatrow}
\end{table}

%% file: Sections/Analysis/04_feature_aggregation.tex
\subsection{Choice of feature aggregation method}
\label{sec:feature_aggregation}

As explained in \cref{sec:taxonomy}, deep learning methods often compute the feature representation of a 3D location by aggregating features from its local neighborhood.
For Bag-of-Points representations, the choice of feature aggregation method often becomes a distinguishing factor, whereas rasterized (2D or 3D) representations are typically used in conjunction with convolutions.
The choice of aggregation method significantly affects the representational capacity and run-time latency.

We observe that each feature aggregation method brings its own trade-off between latency and representational capacity.
The representational capacity of a model is typically correlated with its number of parameters~\cite{liuConvnet2020s2022}.
This is also reflected in \cref{tab:comparison}, as a higher number of parameters is typically correlated with higher performance in semantic segmentation.
Therefore, the ratio between the number of parameters and its run-time latency may be a useful indication of the suitability of an aggregation method for the low-latency use case.\\
In \cref{tab:ops_scaling}, we list the theoretical run-time and parameter scaling of various common feature aggregation methods used in deep neural networks.
Here, $M$ is the number of locations for which features are aggregated, $d$ is the dimensionality of the feature vector for each location, and
$n_\mathrm{neighbors}$ denotes the number of locations in the neighborhood whose features are aggregated.
$t_\mathrm{lookup}$ represents the time required to find a neighboring location in the representation or an external indexing structure, and load its feature vector from memory.
From this table we can observe that convolutions and fully-connected layers have the highest ratio of parameter count to induced latency, especially when using representations with high memory lookup times ($t_\mathrm{lookup}$).
Fully-connected layers do not aggregate features across neighbors. 
As such, they need to be combined with a separate neighborhood aggregation method, which affects the run-time cost of the combination.\\
Another important consideration for selecting a feature aggregation method is the resulting invariance properties of the representation.
For instance, convolutions impart translation invariance into a representation, which might positively affect representational capacity~\cite{triessSemilocalConvolutionsLiDAR2021}.
We encourage future work to experiment with aggregation methods which might provide invariances against point cloud scale, density, or rotation around arbitrary axes.

%% file: Sections/04_Discussion.tex
\section{Discussion}
\label{sec:discussion}
In \cref{sec:analysis}, we examined each design decision in the proposed taxonomy in isolation for their effects on memory consumption, latency, and representational capacity. 
In this section, we take an opposite view point and examine for each outcome -- i.e memory consumption, run-time latency, and representational capacity -- by which design decisions they are most affected.

\paragraph{Memory} The memory consumption of point cloud representations are most prominently affected by the choice of explicit or implicit memory layout.
As listed in \cref{tab:comparison}, the memory consumption characteristics of implicit representations are approximately identical, scaling linearly with the number of points.
Therefore, by choosing an implicit memory layout, any point cloud representation can be reduced to $O(N \cdot d)$ memory complexity, excluding any indexing structures.
Here, $N$ is the number of points in the point cloud, and $d$ is the dimensionality of the feature vector for each point.
As discussed in \cref{sec:explicit_implicit}, representations with an explicit spatial structure are often less memory-efficient due to empty spatial sections being represented in memory.
The memory consumption of rasterized representations is typically dominated by the most fine-grained grid resolution used.
Therefore, it may be useful for certain applications to mix sparse implicit memory layouts for fine-grained rasterizations and use explicit low-resolution feature maps for coarse grid representations, which may be faster to operate on while incurring little sparsity.
As mentioned in \cref{sec:dimensionality} and \cref{sec:cartesian_polar}, some explicit representations incur much less sparsity than others, and therefore can represent the entire point cloud more precisely without much additional memory cost.
Namely, range image based 2D rasterizations can perform their projection in such a way that their memory representation incurs little to no sparsity~\cite{triessScanbasedSemanticSegmentation2020}.
In contrast, Cartesian representations typically incur the highest sparsity, as spatial cells in regions far away from the sensor -- i.e. with low point density -- are still represented by fine-grained grid cells.
Therefore, the choice of coordinate system should also be considered for explicit 2D or 3D rasterizations.

\paragraph{Latency} The run-time latency incurred during the computation of point cloud representations is affected by several design decisions.
As features are aggregated element-wise for each spatial location in the point cloud representation, the number of spatial locations is an important factor.
For implicit spatial representations, this is typically the number of points in the point cloud, whereas for explicit representations it depends on the rasterization resolution.
Given a sufficiently fine spatial resolution, an implicit representation will typically incur far less spatial locations, and thus be much faster to compute.\\
The next factor to consider is the initial latency overhead for the construction of the representation.
This overhead includes the cost of projection, rasterization, but also the cost of constructing additional indexing structures such as KD-trees or hashmaps for implicit representations~\cite{thomasKPConvFlexibleDeformable2019,tangSearchingEfficient3D2020}.
As the overhead is typically unaffected by the scale of the following model, it is particularly important to consider for small models, where it may cause a significant fraction of total incurred latency~\cite{liuPVNAS3DNeural2021}.\\
Another important consideration for latency is the duration of feature aggregation, as discussed in \cref{sec:feature_aggregation}.
This time is affected by the number of neighbors to aggregate, i.e. representation dimensionality and receptive field for aggregation.
However, it is also significantly affected by neighborhood lookup times.
This can be seen particularly for the un-structured Bag-of-Points representations in \cref{tab:comparison}.
Therefore, the choice of explicit vs implicit representation again significantly affects run-time latency.
Finally, the run-time latency also scales with the size of the model used.
As models utilize larger feature dimensionalities, more parameters and more computation, the run-time scales accordingly.
As seen in \cref{tab:comparison}, the scaling of run-time against the number of parameters is highly affected by the choice of specific representation.
As we observe from this comparison, rasterized 2D CNNs typically have some initial overhead, but their latency scales least steeply when increasing their parameter count.
If slightly higher latency is acceptable, sparse 3D CNNs also provide a highly performant alternative~\cite{triessScanbasedSemanticSegmentation2020,liuPVNAS3DNeural2021}.\\
It is important to note that latency is also dependent on the hardware configuration deployed in the vehicle.
This includes both the amount of parallel processing available in GPU-based, FPGA or ASIC hardware for processing, but also includes any ASICs for pre-processing on the sensor's chip itself.
As some sensors provide different "raw" representations in their data output, transformations to particular input representations for a deep learning model may be more or less costly.

\paragraph{Capacity} The representational capacity of point cloud representations is significantly affected by the number of parameters in the model, which can be scaled using the dimensionality $d$ of feature vectors in the representation.
Another important consideration however is how directly the feature aggregation methods can infer the underlying 3D structure from the representation.
As seen in \cref{tab:comparison}, some sparse 3D voxel-based approaches require less parameters to achieve the same segmentation performance as projection-based approaches.
To resolve this issue, some 2D projection-based approaches have added three-dimensional post-processing steps, which can provide improvements to segmentation performance, but also incur run-time latency~\cite{miliotoRangeNetFastAccurate2019}.
Therefore, the choice of representation dimensionality and coordinate system is important for representational capacity.
As discussed in \cref{sec:cartesian_polar}, different coordinate systems also provide different invariances when combined with translation-invariant operations.
Whether polar or Cartesian coordinates are best for representational capacity is still not fully clear.
However, recent results suggest that there may be added benefit from combining polar and Cartesian representations to benefit from invariances to both rotation and translation (see bottom rows in \cshref{tab:comparison})~\cite{xuRPVNetDeepEfficient2021,liCPGNetCascadePointGrid2022}.\\
When using a vehicle setup with multiple LiDAR sensors, one should also carefully consider the coordinate origin and rasterization resolution of the chosen coordinate system. 
This is especially the case for spherical range image projections and other polar coordinate systems.\\
Orthogonal factors such as loss functions and data augmentation have also been shown to bring measurable benefits to representational capacity for little inference time cost~\cite{zhangBagFreebiesTraining2019,triessScanbasedSemanticSegmentation2020,cortinhalSalsaNextFastUncertaintyaware2020,houPointtoVoxelKnowledgeDistillation2022}.
Future work is therefore still required in order to disentangle the representational capacity of point cloud representations from their training methodology.

%
%

%% file: Sections/05_Conclusion.tex
\section{Conclusion}
\label{sec:conclusion}
In this work, we presented a taxonomy covering the most significant design decisions regarding the LiDAR point cloud representation used in deep neural networks.
By categorizing recent state-of-the-art methods for 3D semantic segmentation, we uncover common trends which can be traced back to common design decisions regarding the representational capacity.
We dive deeply into the design decisions made during the development of point cloud representations, and analyze the impact of each design decision on memory consumption, run-time latency and representational capacity.
Finally, we discuss for each of the limitations present in the autonomous vehicle use-case, which design decisions can be altered to lighten the load of latency- or memory-constraints, or increase representational capacity.
The presented taxonomy can be used as a structural basis for future research in the domain of deep learning on point clouds. 
It can further be used to classify different model designs and to reason about their memory consumption, inference time and representational capacity during the design phase of the model.

%% file: Sections/06_Acknowlegement.tex
\begin{ack}
We would like to thank Daniel Bogdoll and Sven Ochs for for their helpful feedback and discussions during the development of this work.

The research leading to these results was conducted within the project KIsSME (Artificial Intelligence for selective near-real-time recordings of scenario and maneuver data in testing highly automated vehicles) and was funded by the German federal ministry for economic affairs and energy.
Responsibility for the information and views set out in this publication lies entirely with the authors.
\end{ack}